\documentclass[runningheads,orivec]{llncs}
\usepackage[T1]{fontenc}

\usepackage{multirow}
\usepackage{adjustbox}
\usepackage{booktabs}
\usepackage{arydshln}
\usepackage{pgfplots}
\usepackage{amsmath}
\usepackage{array}
\usepackage{algorithm2e}
\usepackage{todonotes}
\usepackage[framemethod=TikZ]{mdframed}
\usetikzlibrary{patterns}

\pgfplotsset{compat=1.18}

\definecolor{lightblue}{HTML}{5DA5DA}
\definecolor{lightorange}{HTML}{FAA43A}
\definecolor{lightgreen}{HTML}{60BD68}

\makeatletter
\def\adl@drawiv#1#2#3{%
        \hskip.5\tabcolsep
        \xleaders#3{#2.5\@tempdimb #1{1}#2.5\@tempdimb}%
                #2\z@ plus1fil minus1fil\relax
        \hskip.5\tabcolsep}
\newcommand{\cdashlinelr}[1]{%
  \noalign{\vskip\aboverulesep
           \global\let\@dashdrawstore\adl@draw
           \global\let\adl@draw\adl@drawiv}
  \cdashline{#1}
  \noalign{\global\let\adl@draw\@dashdrawstore
           \vskip\belowrulesep}}
\makeatother

\begin{document}

\title{Large Language Models for the Summarization of Czech Documents: From History to the Present}

\titlerunning{Large Language Models for Summarizing Czech Documents}

\author{V\'{a}clav Tran\inst{1}\orcidID{0009-0003-0250-2821},
Jakub \v{S}m\'id\inst{1, 2}\orcidID{0000-0002-4492-5481},
Ladislav Lenc\inst{1, 2}\orcidID{0000-0002-1066-7269},
Jean-Pierre Salmon\inst{3}\orcidID{0009-0002-0049-7726}
and Pavel Kr\'{a}l\inst{1, 2}\orcidID{0000-0002-3096-675X}}

\authorrunning{V. Tran et al.}
\institute{Department of Computer Science and Engineering, University of West Bohemia in Pilsen, Univerzitn\'{i}, Pilsen, Czech Republic
\and NTIS - New Technologies for the Information Society, University of West Bohemia in Pilsen, Univerzitn\'{i}, Pilsen, Czech Republic \\
\and LaBRI, UMR 5800, F-33400, Univ. de Bordeaux\\
\url{https://nlp.kiv.zcu.cz/}
\\
\email{\{jaksmid,llenc,pkral\}@kiv.zcu.cz,jean-pierre.salmon@u-bordeaux.fr}
}

\maketitle

\begin{abstract}
Text summarization is the task of automatically condensing longer texts into shorter, coherent summaries while preserving the original meaning and key information. Although this task has been extensively studied in English and other high-resource languages, Czech summarization, particularly in the context of historical documents, remains underexplored. This is largely due to the inherent linguistic complexity of Czech and the lack of high-quality annotated datasets.

In this work, we address this gap by leveraging the capabilities of Large Language Models (LLMs), specifically Mistral and mT5, which have demonstrated strong performance across a wide range of natural language processing tasks and multilingual settings. In addition, we also propose a translation-based approach that first translates Czech texts into English, summarizes them using an English-language model, and then translates the summaries back into Czech. Our study makes the following main contributions:
We demonstrate that LLMs achieve new state-of-the-art results on the SumeCzech dataset, a benchmark for modern Czech text summarization, showing the effectiveness of multilingual LLMs even for morphologically rich, medium-resource languages like Czech.
We introduce a new dataset, Posel od \v{C}erchova, designed for the summarization of historical Czech texts. This dataset is derived from digitized 19th-century publications and annotated for abstractive summarization. We provide initial baselines using modern LLMs to facilitate further research in this underrepresented area.

By combining cutting-edge models with both modern and historical Czech datasets, our work lays the foundation for further progress in Czech summarization and contributes valuable resources for future research in Czech historical document processing and low-resource summarization more broadly.
\keywords{Czech Text Summarization \and Deep Neural Networks \and Mistral \and mT5 \and Posel od \v{C}erchova \and SumeCzech \and Transformer Models}
\end{abstract}

\section{Introduction}
\label{sec:introduction}
The rapid advancement of Natural Language Processing (NLP) techniques has substantially improved the capabilities of automated text summarization systems. However, the majority of these developments have been concentrated on high-resource languages such as English, leaving other languages, especially Czech, comparatively underexplored. This disparity becomes even more apparent in the case of historical Czech, which poses unique challenges due to linguistic shifts, archaic vocabulary, and inconsistent syntactic structures.

Effectively summarizing historical Czech texts is not only a technical challenge, but also a task with significant practical value. Many such documents contain culturally and historically important information, often hidden in long and complex narratives. Automated summarization of these texts can greatly support digital humanities, historical research, archival work, and the accessibility of heritage content to broader audiences. By enabling faster navigation and comprehension of large volumes of historical material, summarization tools can accelerate academic workflows and promote the reuse of digitized content in education, journalism, and public history initiatives.

These factors highlight a critical need for the development of robust summarization systems tailored specifically to the domain of historical Czech documents.

To address the current limitations in Czech text summarization, this paper tackles two interrelated challenges. First, we aim to establish new state-of-the-art benchmarks on SumeCzech, the most comprehensive dataset available for modern Czech summarization. We leverage the capabilities of advanced Large Language Models (LLMs), specifically Mistral~\cite{jiang2023mistral} and mT5~\cite{xue-etal-2021-mt5}, to evaluate their effectiveness in handling the linguistic richness and morphological complexity of contemporary Czech.
Moreover, we propose a translation-based approach that first translates Czech texts into English, summarizes them using an English-language model, and then translates the summaries back into Czech.

Then, in light of the evident lack of resources tailored to historical Czech, we introduce a novel dataset constructed from the 19th-century journal Posel od \v{C}erchova. This corpus is carefully curated to support summarization tasks in historical domains, where the language exhibits substantial diachronic variation, including archaic vocabulary and evolving syntactic patterns. The dataset is intended to provide a solid foundation for future research into the summarization of historical Czech texts, where domain-specific challenges differ significantly from those in modern language processing. The full corpus is freely available for research purposes\footnote{https://corpora.kiv.zcu.cz/posel\_od\_cerchova/}, promoting openness and reusability.

Note that this paper presents an extended version of~\cite{icaart25}, including additional experiments, a more detailed methodological description, and a broader discussion of the results and their implications.

By combining methodological advances in LLMs with the creation of a new domain-specific dataset, our work contributes not only to the technical advancement of Czech summarization systems but also to broader applications in cultural heritage preservation, historical scholarship, and the digital humanities. We believe these contributions will help bridge the gap between modern NLP methods and the demands of underrepresented, historically valuable language resources.

\section{Related Work}
Text summarization methods can be categorized into abstractive and extractive ones. Extractive summarization selects the most representative sentences from the source document, while abstractive summarization generates summaries composed of newly created sentences.

Early summarization methods were extractive ones and relied on statistical and graph-based methods like TF-IDF (Term Frequency-Inverse Document Frequency)~\cite{ChristianTFIDFSumm2016}, which scores sentence importance based on term frequency relative to rarity across a corpus.
Similarly, TextRank~\cite{mihalcea-tarau-2004-textrank} represents sentences as nodes in a graph and ranks them using the PageRank algorithm~\cite{Page1999ThePC}.

Neural networks advanced both extractive and also abstractive summarization by modeling sequences with Recurrent Neural Networks (RNNs)~\cite{elman1990finding}. One extractive approach involves sequence-to-sequence architectures where LSTM models capture the contextual importance of each sentence within a document~\cite{nallapati2017summarunner}.
Hierarchical attention networks combine sentence-level and word-level attention to better capture document structure and relevance for summarization~\cite{yang2016hierarchical}. This approach has proven effective in summarizing longer and more complex documents.
Hybrid approaches combining BERT embeddings~\cite{devlin2019bert} with K-Means clustering~\cite{kmeans} to identify key sentences~\cite{miller2019leveraging} have shown excellent performance for abstractive summarization.

Advances in sequence-to-sequence Transformer-based models~\cite{vaswani2017attention} have revolutionized abstractive summarization. Recent models like T5~\cite{raffel2020t5} adopt a text-to-text framework and excel in various tasks, including summarization, due to pre-training on the C4 dataset. PEGASUS~\cite{zhang2019pegasus} introduces gap sentences generation for masking key sentences during pre-training, achieving strong performance on 12 datasets. Similarly, BART~\cite{lewis2019bart} uses denoising objectives for robust text summary generation. Multilingual models such as mT5~\cite{xue-etal-2021-mt5} and mBART~\cite{liu-etal-2020-multilingual-denoising} extend these capabilities to multiple languages, including Czech, through datasets like mC4~\cite{xue-etal-2021-mc4} and multilingual Common Crawl\footnote{http://commoncrawl.org/}.

However, these models often underperform on non-English corpora without fine-tuning.

\section{Datasets}
The following section provides a brief review of the primary existing summarization datasets.
Moreover, the created {Posel od \v{C}erchova} corpus will also be detailed at the end of this section.

\subsection{English Datasets}
{\bf CNN/Daily Mail} dataset~\cite{hermann2015cnndm} comprises more than 300,000 English-language news articles, each accompanied by highlights written by the article authors. This extensive dataset has become a cornerstone in the field of natural language processing, particularly in research related to abstractive summarization and question-answering systems. Over time, it has evolved through multiple versions, with adaptations designed to address specific NLP tasks and to improve the quality and diversity of the training data.
\\ \\
\noindent
{\bf XSum} dataset~\cite{xsumDataset} consists of 226,000 BBC articles paired with concise, single-sentence summaries. These articles span a wide range of domains, including news, sports, and science, providing a diverse and comprehensive dataset for summarization tasks. By emphasizing one-sentence summaries, XSum is specifically designed to challenge and promote the development of abstractive summarization models, which generate concise, human-like summaries rather than relying heavily on extractive methods. 
\\ \\
\noindent
{\bf ArXiv} Dataset~\cite{cohan-etal-2018-discourse} comprises 215,000 pairs of scientific papers and their corresponding abstracts, all sourced from the arXiv preprint repository. This dataset has undergone extensive cleaning and standardization to facilitate research and analysis, which includes the removal of non-essential sections such as figures, tables, and references to focus solely on the textual content. 
\\ \\
\noindent
{\bf BOOKSUM}~\cite{kryscinski-etal-2022-booksum} is a dataset tailored for summarizing long texts like novels, plays, and stories, with summaries provided at paragraph, chapter, and book levels. Texts and summaries were sourced from Project Gutenberg and other web archives, supporting both extractive and abstractive summarization.

\subsection{Multilingual Datasets}
{\bf XLSum}~\cite{xlsumDataset} provides over one million article-summary pairs across 44 languages, ranging from low-resource languages like Bengali and Swahili to high-resource languages such as English and Russian. Extracted from various BBC sites, this dataset is a valuable resource for multilingual summarization research.
\\ \\
\noindent
{\bf MLSUM}~\cite{mlsumDataset} consists of 1.5 million article-summary pairs in five languages: German, Russian, French, Spanish, and Turkish. The dataset was created by archiving news articles from well-known newspapers, including Le Monde and El Pais, with a focus on ensuring broad topic coverage.

While the above datasets represent significant progress in summarization research, especially for English and several other widely spoken languages, the availability of comparable resources for Czech remains extremely limited. In particular, Czech lacks large-scale, high-quality datasets for both modern and historical domains. This scarcity significantly hinders the development and evaluation of summarization models tailored to Czech, and motivates the need for focused dataset creation and model adaptation in this underrepresented linguistic setting.

\subsection{SumeCzech}
SumeCzech is a large-scale dataset for Czech summarization~\cite{straka2018sumeczech} and represents a notable exception to the general scarcity of Czech-specific NLP resources. Created by the Institute of Formal and Applied Linguistics at Charles University, it is specifically designed to support summarization tasks in the Czech language.

The dataset consists of approximately one million news articles collected from five major Czech news outlets: České Noviny, Deník, iDNES, Lidovky, and Novinky.cz. Each entry is stored in JSONLines format and includes rich metadata such as the article URL, headline, abstract, full text, subdomain, section, and publication date. The creators applied preprocessing steps including language detection, duplicate removal, and filtering out records with empty or excessively short fields to ensure data quality.

SumeCzech is well-suited for multiple summarization tasks, including both headline generation and multi-sentence abstract generation, which makes it a versatile resource for training and evaluating models. The dataset is split into training, development, and test sets in roughly an 86.5/4.5/4.5 ratio. On average, full texts contain around 409 words, while the associated abstracts average 38 words, providing a reasonable length ratio for abstractive summarization.

Despite its scale and utility, SumeCzech is focused entirely on modern Czech language and content, limiting its applicability to historical text processing, which presents different linguistic characteristics such as archaic spelling, grammar, and vocabulary. This gap underscores the need for dedicated datasets targeting historical Czech, especially for tasks like summarization that rely on linguistic fluency and domain-specific adaptation.

\subsection{Posel od \v{C}erchova}
 To construct the dataset, we used data from the historical journal \textit{Posel od \v{C}erchova (POC)}, which is available on the archival portal Porta fontium\footnote{\url{https://www.portafontium.eu}}.
 
The construction of the dataset involved addressing the challenge of creating summaries for the provided texts, which were composed in historical Czech and, in some rare cases, even German. The texts also covered a variety of different topics, from local news surrounding Domažlice (a historic town in the Czech Republic), opinion pieces, and various local advertisements to internal and worldwide politics and feuilletons. Furthermore, it was important to create a dataset of sufficient size to ensure the accuracy and reliability of the evaluation. These aspects added complexity to the summarization task.

To overcome the mentioned issues, we employed state-of-the-art (SOTA) LLMs GPT-4~\cite{openai2024gpt4} and Claude 3 Opus~\cite{anthropic-2024} (Opus) (specifically the \texttt{claude-3-opus- 20240229} version) for initial text summary creation. These models were selected based on their SOTA performance in many NLP tasks and excellent performance in some preliminary summarization experiments.

While generating the summaries, it was essential to ensure conciseness. Since most of the implemented methods were fine-tuned on the SumeCzech dataset, we aimed to maintain consistency by creating summaries in a journalistic style, reflecting the dataset's characteristics. To achieve this, the prompts for generating the summaries included explicit instructions, as shown below:

\begin{itemize}
\item Vytvoř shrnutí následujícího textu ve stylu novináře. Počet vět $<= 5$; (EN: Create a summary of the following text in the style of a journalist. Number of sentences $<= 5$)
\end{itemize}

During the summarization task, we observed that while both models produced summaries of very good quality, Opus tended to create more succinct and stylistically appropriate ones, closely aligning with the news reporter format. However, there were instances where summaries generated by Opus exhibited an excessive focus on a single topic.

On the other hand, GPT-4 aimed to incorporate a greater level of detail within the five-sentence constraint but occasionally deviated from the specified stylistic prompt.

If the model-generated summary exhibited significant stylistic deviations or excessive focus on a single topic, we either modified or regenerated it until a correct version was achieved.

Two levels of summaries were created: the first at the page level, and the second summarizing an entire article, which typically consists of several pages. In total, we summarized 432 pages, which effectively resulted in the creation of 100 article-level summaries. The subset containing the page-level summaries is hereafter referred to as {\it POC-P}, while the article-level summaries are referred to as {\it POC-I}.
Note that all created summaries were checked and corrected manually by two native Czech speakers.

The dataset is in the {\bf .json} format and contains the following information:

\begin{itemize}
    \item \textbf{text:} Text extracted from the given page, a digital rendition of the original printed content;
    \item \textbf{summary:} Summary of the page, which is no more than 5 sentences long;
    \item \textbf{year:} Publication year of the journal;
    \item \textbf{journal:} Specification of the source journal: the day, month, and the number of the issue is contained within this identifier;
    \item \textbf{page\_src:} Name of the source image file converted into the text;
    \item \textbf{page\_num:} Page number.
\end{itemize}

This dataset is designed to support summarization tasks within Czech historical contexts, providing researchers with the tools to tackle the linguistic challenges unique to this domain. The corpus is freely available for research purposes\footnote{https://corpora.kiv.zcu.cz/posel\_od\_cerchova/}.

\section{Methods}
The experiments employ two advanced Transformer-based models, Multilingual Text-to-Text Transfer Transformer (mT5)~\cite{xue-etal-2021-mt5} and Mistral 7B~\cite{jiang2023mistral}.

Moreover, we propose an alternative approach to abstractive summarization that translates the Czech text into English, summarizes it using an English model, and then translates it back, hereafter referred to as the {\it Translation-Summarization-Translation (TST)} method.

\subsection{Multilingual Text-to-Text Transfer Transformer}
The Multilingual Text-to-Text Transfer Transformer (mT5) is a variant of the T5 model designed for multilingual tasks. This model is trained on the multilingual mC4 dataset~\cite{xue-etal-2021-mc4}, which includes Czech, and effectively handles a wide range of languages. The model is based on Transformer encoder-decoder architecture and uses a SentencePiece tokenizer~\cite{kudo-richardson-2018-sentencepiece} to process complex language structures, including Czech morphology. Pre-trained using a span corruption objective~\cite{raffel2020t5}, mT5 predicts masked spans of text, enabling it to learn semantic and contextual relationships.

The mT5 model is available in various sizes, from small with 300 million parameters to XXL with 13 billion parameters, and is therefore adapted to different computational needs. The base variant of the mT5, which contains 580 million parameters, is used for further experiments.

\subsection{Mistral Language Model}
The Mistral Language Model (Mistral LM) is a highly efficient large language model known for its robust performance across diverse natural language processing tasks. It is designed to combine high accuracy with computational efficiency, achieving state-of-the-art results in reasoning, text generation, summarization, and other NLP applications. Mistral 7B, with its 7 billion parameters, strikes a balance between computational efficiency and task performance, surpassing larger models like 13B or 34B in several benchmarks.

This model utilizes advanced attention mechanisms like Grouped-Query Attention (GQA)~\cite{ainslie2023gqa} and Sliding Window Attention (SWA)~\cite{beltagy2020longformer}. GQA enhances processing speed by grouping attention heads to focus on the same input data, while SWA reduces computational costs by limiting token attention to nearby tokens. The model supports techniques such as quantization~\cite{gholami2021survey} and Low-Rank Adaptation (LoRA)~\cite{hu2021lora} for efficient fine-tuning on limited hardware, enabling it to handle longer inputs effectively.

\subsection{Translation-Summarization-Translation (TST) Method}
As already stated, this approach involves translating the Czech text into English, summarizing it using an English-language model, and then translating the summary back into Czech.
For translation, we use ALMA-R model, introduced by Xu et al.~\cite{xu2024contrastive}\footnote{\url{https://github.com/fe1ixxu/ALMA}} which  exceeds the performance of GPT-4~\cite{openai2024gpt4} and WMT\footnote{\url{https://machinetranslate.org/wmt}}, winners on various translation benchmarks. 

The model was trained using {Contrastive Preference Optimization (CPO)}, a novel training method that was proposed in the paper above. Traditional methods such as {Supervised Fine-Tuning (SFT)}~\cite{goodfellow2016deep} train models to mimic reference translations, where the resulting model performance relies on dataset quality. Xu et al. demonstrate that translations by advanced models can be superior to reference translations. Its training objective is designed not to minimize the error between the model output and a single reference translation, but rather to increase the likelihood of generating a preferred translation and decrease the likelihood of generating a dispreferred one. This involves generating a triplet of translations for a given source sentence: one from a reference (human-generated), one from GPT-4, and one from an ALMA~\cite{xu2024paradigm} model (prior to CPO application). Each translation in the triplet is then scored using reference-free translation quality evaluation models and the translations are ranked based on their quality. The highest-scoring translation is labeled as the preferred translation.
\section{Experiments}

\subsection{Evaluation Metrics}
The following evaluation metrics are used.

ROUGE (Recall-Oriented Understudy for Gisting Evaluation)~\cite{lin-2004-rouge} is a set of metrics used to evaluate the quality of summaries by comparing n-gram overlaps between a system-generated summary and reference texts. Key ROUGE metrics include ROUGE-N (for n-gram overlap) and ROUGE-L (for the longest common subsequence).

ROUGERAW~\cite{straka2018rougeraw} is a variant of ROUGE that evaluates raw token-level overlaps between predicted and reference texts without any preprocessing like stemming or lemmatization. It measures exact matches of tokens, making it suitable for tasks where precise token alignment is important.

\subsection{Experimental Setup}
 We used AdamW optimizer~\cite{loshchilov2017decoupled} with a learning rate set to 0.001 as suggested by authors of mT5~\cite{xue-etal-2021-mt5} for the training of this model.
For Mistral 7B, we utilized QLoRA~\cite{qlora}, a method that integrates a 4-bit quantized model with a small, newly introduced set of learnable parameters. During fine-tuning, only these additional parameters are updated while the original model remains frozen, thereby substantially reducing memory requirements.

We used the models from the HuggingFace Transformers library~\cite{wolf-etal-2020-transformers}.
 For training both models, we used a single NVIDIA A40 GPU with 45 GB VRAM.

\subsection{Model Variants}
We use the following model variants in our experiments:
\begin{itemize}
    \item M7B-SC: The Mistral 7B model fine-tuned on the SumeCzech dataset;
    \item M7B-POC: The Mistral 7B model further fine-tuned on the POC dataset;
    \item mT5-SC: The mT5 model fine-tuned on the SumeCzech dataset;
    \item TST: used ALMA-R 13B model for translation, English summarization was done using the 4-bit pre-quantized, fine-tuned variant of Mistral 7B\footnote{\url{https://huggingface.co/unsloth/mistral-7b-instruct-v0.2-bnb-4bit}}.
\end{itemize}

\subsection{Results on the SumeCzech Dataset}

This experiment compares the results of the proposed mT5-SC and M7B-SC models with related work on the SumeCzech dataset, see Table~\ref{tab:eval:sc_scores_comparison}.

The first comparative method, HT2A-S~\cite{Krotil2022TextSummarization}, is based on the mBART model, which is further fine-tuned on the SumeCzech dataset.
The other methods provided by the authors of the SumeCzech dataset~\cite{straka2018sumeczech} are as follows:  First, Random, Textrank and Tensor2Tensor~\cite{tensor2tensor}.

Table~\ref{tab:eval:sc_scores_comparison} demonstrates that the proposed M7B-SC method is very efficient, outperforming all other baselines and achieving new state-of-the-art results on this dataset.
Furthermore, the second proposed approach, mT5-SC, also performs remarkably well, consistently obtaining the second-best results.
 
 \begin{table*}[ht!]
    \centering
    \caption{Results of various methods on SumeCzech dataset with precision (P), recall (R), and F1-score (F)~\cite{icaart25}.}
    \label{tab:eval:sc_scores_comparison}
    \begin{tabular}{
        l
        r
        r
        r
        r
        r
        r
        r
        r
        r
    }
        \toprule
        Method & \multicolumn{3}{c}{ROUGE\textsubscript{raw}-1} & \multicolumn{3}{c}{ROUGE\textsubscript{raw}-2} & \multicolumn{3}{c}{ROUGE\textsubscript{raw}-L} \\
        \cmidrule(lr){2-4} \cmidrule(lr){5-7} \cmidrule(lr){8-10}
        & {P} & {R} & {F} & {P} & {R} & {F} & {P} & {R} & {F} \\
        \midrule
        M7B-SC & \textbf{24.4} & \textbf{19.7} & \textbf{21.2} & \textbf{6.5} & \textbf{5.3} & \textbf{5.7} & \textbf{17.8} & \textbf{14.5} & \textbf{15.5} \\
        mT5-SC & 22.0 & 17.9 & 19.2 & 5.3 & 4.3 & 4.6 & 16.1 & 13.2 & 14.1 \\
        \midrule
        HT2A-S~\cite{Krotil2022TextSummarization} & 22.9 & 16.0 & 18.2 & 5.7 & 4.0 & 4.6 & 16.9 & 11.9 & 13.5 \\
        First~\cite{straka2018sumeczech} & 13.1 & 17.9 & 14.4 & 0.1 & 9.8 & 0.2 & 1.1 & 8.8 & 0.9 \\
        Random~\cite{straka2018sumeczech} & 11.7 & 15.5 & 12.7 & 0.1 & 2.0 & 0.1 & 0.7 & 10.3 & 0.8 \\
        Textrank~\cite{straka2018sumeczech} & 11.1 & 20.8 & 13.8 & 0.1 & 6.0 & 0.3 & 0.7 & 13.4 & 0.8 \\
        Tensor2Tensor~\cite{straka2018sumeczech} & 13.2 & 10.5 & 11.3 & 0.1 & 2.0 & 0.1 & 0.2 & 8.1 & 0.8 \\
        \bottomrule
    \end{tabular}
\end{table*}

\subsection{Results on Posel od \v{C}erchova Dataset}
This section evaluates the proposed methods on the Posel od \v{C}erchova dataset. Table~\ref{tab:eval:poc-p} shows the results on the {\it POC-P} subset containing summaries for every page (106 pages), while Table~\ref{tab:eval:poc-i} depicts the results on the {\it POC-I} subset, which is composed of the summaries of every article (25 issues).

These tables show that, as in the previous case, M7T-POC model gives significantly better results than the mT5-SC model, and it is with a very high margin. 
Additionally, the TST approach slightly outperformed the M7T-POC model in some cases.

\begin{table*}[ht!]
    \centering
    \caption{Results of implemented methods on the {\it POC-P} subset from Posel od \v{C}erchova dataset with precision (P), recall (R), and F1-score (F).}
    \label{tab:eval:poc-p}
    \begin{tabular}{
        l
        r
        r
        r
        r
        r
        r
        r
        r
        r
    }
        \toprule
        Method & \multicolumn{3}{c}{ROUGE\textsubscript{raw}-1} & \multicolumn{3}{c}{ROUGE\textsubscript{raw}-2} & \multicolumn{3}{c}{ROUGE\textsubscript{raw}-L} \\
        \cmidrule(lr){2-4} \cmidrule(lr){5-7} \cmidrule(lr){8-10}
        & {P} & {R} & {F} & {P} & {R} & {F} & {P} & {R} & {F} \\
        \midrule
        M7B-POC & \textbf{23.5} & {17.4} & {19.6} & \textbf{4.8} & {3.5} & \textbf{4.0} & \textbf{16.6} & {12.2} & \textbf{13.8} \\
        mT5-SC & 20.2 & 8.2 & 11.1 & 1.4 & 0.5 & 0.7 & 14.9 & 6.1 & 8.2 \\
        TST & 17.2 & \textbf{25.1} & \textbf{19.9} & 2.5 & \textbf{3.8} & 2.9 & 11.3 & \textbf{16.4} & 13.0 \\
        \bottomrule
    \end{tabular}
\end{table*}

\begin{table*}[ht!]
    \centering
    \caption{Results of implemented methods on {\it POC-I} subset from Posel od \v{C}erchova dataset with precision (P), recall (R), and F1-score (F).}
    \label{tab:eval:poc-i}
    \begin{tabular}{
        l
        r
        r
        r
        r
        r
        r
        r
        r
        r
    }
        \toprule
        Method & \multicolumn{3}{c}{ROUGE\textsubscript{raw}-1} & \multicolumn{3}{c}{ROUGE\textsubscript{raw}-2} & \multicolumn{3}{c}{ROUGE\textsubscript{raw}-L} \\
        \cmidrule(lr){2-4} \cmidrule(lr){5-7} \cmidrule(lr){8-10}
        & {P} & {R} & {F} & {P} & {R} & {F} & {P} & {R} & {F} \\
        \midrule
        M7B-POC & \textbf{19.3} & {17.6} & \textbf{18.0} & \textbf{3.2} & {2.8} & \textbf{2.9} & { 13.7} & {12.4} & \textbf{12.8} \\
        mT5-SC & 18.2 & 5.9 & 8.6 & 1.0 & 0.3 & 0.4 & \textbf{14.0} & 4.5 & 6.5 \\
        TST & 14.0 & \textbf{24.8} & 17.5 & 1.7 & \textbf{3.1} & 2.1 & 9.1 & \textbf{16.3} & 11.4 \\
        \bottomrule
    \end{tabular}
\end{table*}

\section{Discussion}
We provide two examples of generated summaries. The first one, depicted in Table \ref{eval:tab:example1}, presents page-level summaries from the 38th issue of Posel od Čerchova (1882). The second example, shown in Table \ref{eval:tab:example2}, contains issue-level summaries from the 52nd issue of the same publication and year.

The summaries in Table \ref{eval:tab:example1} generally convey similar information as the reference summary to a certain extent. However, the mT5-SC summary is just a composition of the first two sentences of the original text, which is not shown here due to its length. Both the M7B-POC and TST summaries contain grammatical errors. While the M7B-POC summary is slightly more coherent and better structured, it also includes more factual inaccuracies, as evident in its concluding sentence, where it mistakenly mentions the city of Prague as one of the donors and attributes the Weihburggasse street to the wrong village.

The following example in Table \ref{eval:tab:example2} shows comparable results. The mT5-SC summary was irrelevant. The M7B-POC summary has slightly better coherence and grammar in comparison to TST summary, however, it misinforms by not stating that the death of "sultánka" is a fictional event.

In our limited observations of these generated summaries and many other generated summaries, M7B-POC tended to create slightly more coherent summaries with better grammar, but TST produced less factual errors than M7B-POC.
\begin{table}[h!] 
\centering
\caption{Summaries of 'Posel od Čerchova,' page 5 of the 38th issue, 1882}
\label{eval:tab:example1}
\begin{tabular}{@{}p{0.95\textwidth}@{}} 
\toprule
\textbf{Reference} \\
\midrule
{\small V Petrohradě došlo k otravě osmi běloušů, kteří měli táhnout korunovační vůz cara Alexandra III. při jeho korunovaci v Moskvě. Koně byli nalezeni mrtví ve stáji carského paláce. Obce Milaveč a Radonice v okresu Domažlice přispěly na znovuzřízení Národního divadla, čímž mohou jít příkladem ostatnímu venkovu. Článek dále uvádí tržní ceny různých produktů v Domažlicích a Praze.} \\
\midrule
\textbf{M7B-POC} \\
{\small V Petrohradě byli otráveni osm carští bělouši, kteří měli táhnouti korunovační vozku pro cara Alexandra III. v Moskvě. V Domažlicích byly zaznamenány tržní ceny potravin a dalších zboží, včetně cen za talíře a sklenice. V Praze a Weihburggasse 26 v Milaveči a Radonicích I. byly oznámeny příspěvky na obnovu Národního divadla.} \\
\midrule
\textbf{mT5-SC} \\
{\small Carští koně otráveni. Oněch osm běloušů, kteří měli korunní vůz při korunovací carovi Alexandra III. v Moskvě táhnouti, nalezli služebníci ve stáji paláce carského v Petrohradě zdatná ležeti Dla -x-1 byly snad jediné obce v okresu Domažlice.} \\
\midrule
\textbf{TST} \\
{\small V Petrohradě byly v koníčkách palácových stájí nalezeny osm bílých koní, kteří zemřeli na otravu, určených pro korunovaci cara Alexandra III. V českém okrese Domažlice se loni dvě obce podílely na obnově Národního divadla a další mohou následovat jejich příklad. V následujícím textu jsou uvedeny ceny různých položek v Praze: Neffův obchod na Staroměstském náměstí za 39 J. Milaveč a Radonice I. v ulici Weihburggasse 26 vybraly za měsíc 10 zl. díky úsilí několika odvážných občanů.} \\
\bottomrule
\end{tabular}
\end{table}

\begin{table}[h!] 
\centering
\caption{Summaries of 'Posel od Čerchova,' 52nd issue, 1882}
\label{eval:tab:example2}
\begin{tabular}{@{}p{0.95\textwidth}@{}} 
\toprule
\textbf{Reference} \\
\midrule
{\small Domažlický týdeník "Posel od Čerchova" vstupuje do 11. ročníku svého vydávání s čtvrtletním předplatným variujícím dle způsobu doručení. V aktuálním vydání týdeník pokrývá široké spektrum témat, od politických událostí v Bosně, Tyrolsku, až po tragické incidenty v Srbsku a soudní případy v Paříži. Součástí obsahu je také pokračování povídky "Ze života sultánky", odhalení pomníku knížete Michajla v Bělehradě, a záměr francouzského ministra války rozšířit armádu. Zajímavostí jsou novinky na trhu, jako Hellerovy hrací stroje, a opatření pražské městské rady proti novoroční žebrotnické praxi mezi městskými zaměstnanci.} \\
\midrule
\textbf{M7B-POC} \\
{\small Posel od Čerchova informuje o politických událostech, jako jsou předlohy o výpomoci v Tyrolsku a Korutanech, rakousko-německém napětí a ruské vojenské aktivitě. Dále se diskutuje o stavebních pracích v Tyrolsku a Korutanech, financovaných zemskou zárukou. Zmíněna je nová předloha o založení samostatného církevního systému v Sarajevu, umožňující větší autonomii muslimské komunity. Dále je zmíněna popisná smrt sultánky ze Cařihradu, jejího vyhnanství po milostném vztahu s bratrem, a slavnostní odhalení pomníku knížeti Miloše Obrenoviče v Bělehradě. Dále jsou zmíněny kulturní události, jako je nová výstava uměleckých prací v Plzni, a společenské události, jako je nová výzva do sněmu v Bělehradě, a trestní události, jako jsou vraždy a zatčení v několika městech.} \\
\midrule
\textbf{mT5-SC} \\
{\small Vánoční svátky jsou v plném proudu a s nimi i přípravy na vánoční svátky. Přinášíme vám plné znění projevu prezidenta republiky Václava Klause, který přednesl ve Vladislavském sále Pražského hradu.} \\
\midrule
\textbf{TST} \\
{\small ČERCHOVSKÝ TÝDENÍK (POŠLA OD ČERCHOVA), noviny z Domažlic, pokrývají politická, obecná a komunitní témata v daném regionu. Ceny předplatného se liší. V Bosně byla založena náboženská organizace pro bosenskou muslimskou církev, zatímco ve městě Krivošiji byly z důvodu počasí pozastaveny stavební práce. V Srbsku zůstává Královský sněm krále Milana prázdný a situace je po volbách, jejichž cílem bylo nahradit radikální poslance, klidná. V Egyptě jsou odsouzení vůdci arabského povstání veřejně zesměšňováni a poté vysídleni, v České republice došlo k fatálnímu střetu mezi sousedy. Probíhají různé akce, například oslavy jmenování pana Justina V. Práška na funkci ředitele, jmenování paní Karol Fremuthové do funkce a hudební zkoušky pro taneční zábavy 26. prosince. Dále se zde koná loterie ve prospěch organizace „Matica“ a je zveřejněna kniha „Humory“. Společnost Singer nabízí originální šicí stroje jako ideální vánoční dárek.} \\
\bottomrule
\end{tabular}
\end{table}

\section{Conclusions}
\label{sec:conclusion}
This paper investigates the use of state-of-the-art large language models—specifically Mistral 7B and mT5—for Czech text summarization, addressing both contemporary and historical domains. In addition to direct summarization, we propose a novel translation-based pipeline, which first translates Czech input texts into English, performs summarization using an English-language model, and subsequently translates the summaries back into Czech. This method leverages the strength of English-language models in low-resource scenarios.

Our experimental results demonstrate that the proposed Mistral-based model (M7B-SC) sets a new performance benchmark on the SumeCzech dataset, achieving state-of-the-art results. The mT5-based model (mT5-SC) also delivered consistently strong performance, ranking second across most metrics. Moreover, the translation-based approach (TST) proved to be highly competitive, further validating its potential as a viable strategy for Czech summarization tasks.

To support historical text summarization, we introduced Posel od \v{C}erchova, a newly curated dataset of historical Czech documents. Alongside this resource, we provide baseline results and discuss the linguistic and structural challenges that distinguish historical Czech from modern usage.

These contributions not only advance the field of Czech Together, these contributions significantly advance Czech-language summarization research. They also establish a foundation for broader investigations in the processing of historical texts, with practical applications in cultural heritage preservation and digital humanities. Future work may explore improving model robustness, developing hybrid architectures, and expanding the dataset to enable multilingual and diachronic studies.

\begin{credits}
\subsubsection{\ackname} 
The work of Jakub \v{S}m\'{i}d has been supported by the Grant No. SGS-2025-022 -- New Data Processing Methods in Current Areas of Computer Science.
The work of Ladislav Lenc and Pavel Kr\'al has been supported by the project R\&D of Technologies for Advanced Digitalization in the Pilsen Metropolitan Area (DigiTech) No. CZ.02.01.01/00/23\_021/0008436. 
Computational resources were provided by the e-INFRA CZ project (ID:90254), supported by the Ministry of Education, Youth and Sports of the Czech Republic.
Computational resources were provided by the e-INFRA CZ project (ID:90254), supported by the Ministry of Education, Youth and Sports of the Czech Republic.

\subsubsection{\discintname}
The authors have no competing interests to declare that are relevant to the content of this article.
\end{credits}

\bibliographystyle{splncs04}
\bibliography{references}

\end{document}